# Explainable and Adaptive Machine Learning for Fraud Risk Detection in Bangladesh's Digital Banking Ecosystem

Md Najmus Swaqeeb

Designation: Independent Researcher (System Analyst); BRAC University Alumnus

BRAC University, Dhaka, Bangladesh

Phone/Mobile: +880 1882118855

Email: n.swaqeeb@gmail.com | ORCID: https://orcid.org/0009-0001-0355-8064

**Abstract.** The rapid expansion of digital banking and mobile financial services (MFS) in Bangladesh has increased financial inclusion while creating greater exposure to fraud, identity-related risks, and cybersecurity threats. Bangladesh Bank's updated 2026 electronic Know Your Customer (e-KYC) guidelines recognize the growth of technology-based banking and digital payments alongside concerns regarding authenticity and fraudulent activities (Bangladesh Bank, 2026). Recent Bangladesh-focused studies have explored machine-learning-based mobile financial fraud detection and artificial intelligence applications for e-KYC and identity-fraud prevention, highlighting the limitations of traditional and rule-based approaches and the need for adaptive, transparent, and context-specific solutions (Sarker & Tanvir, 2025). This study proposes an explainable and adaptive machine-learning framework for transaction-level fraud-risk detection in Bangladesh's digital banking ecosystem. The framework integrates transaction and behavioral characteristics with supervised machine-learning and anomaly-detection techniques to identify potentially fraudulent activities under highly imbalanced data conditions. Model performance will be evaluated using precision, recall, F1-score, precision-recall area under the curve (PR-AUC), and false-positive rate, with particular attention to the operational consequences of incorrect fraud alerts. Explainable artificial intelligence (XAI) will be incorporated to identify the key factors contributing to individual fraud-risk predictions, thereby improving transparency and supporting human decision-making. An adaptive risk-scoring mechanism will further address changes in transaction behavior and evolving fraud patterns. Building on evidence that AI can strengthen fraud detection while raising important concerns about privacy and explainability in mobile financial services (Mollik & Majeed, 2025), the study aims to provide a practical AI-based approach for strengthening fraud monitoring, customer protection, and ICT risk management in Bangladesh's digital banking sector.

## 1. Introduction

The expansion of digital banking and mobile financial services has increased the importance of effective methods for monitoring potentially fraudulent transactions. Bangladesh provides a relevant application context because its financial sector is increasingly shaped by technology-enabled banking, digital payments, and electronic identity verification. Bangladesh Bank's 2026 e-KYC guidance reflects this broader regulatory attention to technology-based financial processes and associated authenticity and fraud risks (Bangladesh Bank, 2026). Bangladesh-focused studies have also explored machine-learning approaches to mobile financial fraud detection and related identity-risk problems (Sarker & Tanvir, 2025).

Machine-learning methods can model nonlinear relationships among transaction characteristics and can be evaluated against conventional statistical baselines. Random Forest aggregates decision trees to improve predictive performance and robustness (Breiman, 2001), while gradient-boosting frameworks such as XGBoost and LightGBM provide efficient approaches for learning complex patterns from structured data (Chen & Guestrin, 2016; Ke et al., 2017). In fraud detection, however, overall accuracy alone can be misleading because fraudulent transactions usually represent a small proportion of observations. Precision, recall, F1-score, PR-AUC, ROC-AUC, and false-positive rate therefore provide complementary views of performance.

A second issue is the interpretation of model outputs. High predictive performance does not by itself explain why a transaction receives a high fraud-risk score. SHAP provides a feature-attribution framework that can explain model predictions globally and locally (Lundberg & Lee, 2017). A third issue is probability reliability: a model may rank observations effectively while its numerical scores are not well calibrated. Platt scaling is a widely used post-processing approach for calibrating classifier outputs (Platt, 1999).

This study develops an integrated experimental framework combining predictive modeling, probability calibration, explainability, and adaptive risk prioritization. The empirical evaluation uses PaySim, a publicly available synthetic mobile-money dataset developed for fraud-detection research (Lopez-Rojas et al., 2016). Bangladesh is the application context of the study; however, PaySim is not a dataset of Bangladesh banking or MFS transactions. Accordingly, the experimental findings are presented as evidence about the evaluated methodology and benchmark, not as direct estimates of fraud prevalence, fraud behavior, or deployed-model performance in Bangladesh.

The study makes four contributions. First, it applies chronological data partitioning and historical features constructed from preceding observations only. Second, it compares Logistic Regression, Random Forest, Clean XGBoost, and LightGBM using both ranking and threshold-dependent metrics. Third, it evaluates Platt calibration and SHAP explanations alongside predictive performance. Fourth, it compares a fixed-threshold policy with an adaptive top-1% time-window risk-prioritization policy to make the false-positive/recall trade-off explicit.

A useful fraud-detection system should identify a meaningful proportion of suspicious activity while avoiding an alert volume that overwhelms investigators or unnecessarily affects legitimate customers. This is particularly important under severe class imbalance. A classifier can obtain high overall accuracy by predicting the majority class while failing to identify the minority class effectively. The present study therefore evaluates several complementary measures rather than treating accuracy as the primary criterion. Precision describes the proportion of alerts that are fraudulent, recall describes the proportion of observed fraud that is detected, and F1-score summarizes the balance between precision and recall. PR-AUC and ROC-AUC evaluate ranking behavior across thresholds, while false-positive rate provides a direct measure of unnecessary alerts among legitimate transactions.

Fraud-risk detection is also inherently temporal. Transactions occur in an ordered stream, and information generated after a transaction should not be used to construct a feature intended to represent the information available when that transaction is assessed. Randomly mixing observations from different periods can therefore make future performance appear stronger than it may be under a sequential deployment setting. The present design addresses this issue by using chronological partitions and historical features constructed from preceding observations. This does not make the benchmark equivalent to a production banking environment, but it provides a more temporally coherent experimental protocol.

The study also distinguishes discrimination, probability calibration, explainability, and decision policy. A model can rank transactions effectively while producing probabilities that are poorly aligned with observed outcomes. Conversely, a calibrated model may provide more meaningful numerical risk scores without maximizing a threshold-dependent metric such as F1. Similarly, a fixed threshold and a capacity-constrained ranking policy can produce different numbers of alerts and different types of errors. Treating these as separate but connected evaluation questions allows the completed experiments to show not only which model performed best, but also how its outputs behave when converted into operational risk decisions.

## 2. Literature Review

Machine learning has been widely applied to fraud and anomaly detection because transaction data can contain nonlinear interactions and highly imbalanced outcomes. Random Forest provides a standard tree-ensemble baseline (Breiman, 2001), while XGBoost and LightGBM are established gradient-boosting frameworks for structured prediction problems (Chen & Guestrin, 2016; Ke et al., 2017). Prior work using mobile-payment data has also emphasized the importance of evaluating class imbalance and operational consequences rather than relying only on accuracy (Hajek et al., 2023).

A recurring challenge is limited access to real financial transaction data. PaySim was developed as a simulator for mobile-money transactions and has become a reproducible benchmark for fraud-detection research (Lopez-Rojas et al., 2016). Its public availability supports controlled comparison of methods, but its synthetic nature also limits direct generalization to a particular financial system.

Recent work has increasingly incorporated explainable AI into fraud modeling. SHAP-based analysis can provide both global feature importance and transaction-level explanations (Lundberg & Lee, 2017). Probability calibration is also relevant when scores are used for prioritization; Platt scaling transforms model outputs through a fitted logistic model (Platt, 1999).

Bangladesh-focused research provides motivation for investigating adaptive and transparent approaches in the local digital-finance context (Sarker & Tanvir, 2025). The present study does not claim to validate a production system in Bangladesh. Instead, it evaluates a reproducible pipeline that jointly considers discrimination, calibration, explanation, and alternative alert-selection policies.

In this setting, the usefulness of a model depends on its ability to identify minority-class observations without generating excessive false alerts. Research using transaction and mobile-payment data therefore commonly considers precision, recall, F1-score, and area-under-curve measures alongside conventional classification statistics. The present study follows this multi-metric perspective and additionally reports confusion-matrix counts so that the practical scale of false positives and true positives remains visible.

Ensemble tree methods are attractive for structured transaction data because they can capture nonlinear relationships and interactions without requiring a manually specified linear form. Random Forest aggregates multiple decision trees, while gradient boosting sequentially combines weak learners to improve predictive performance. XGBoost and LightGBM are established boosting implementations for structured data. Including Logistic Regression alongside these models provides a comparatively simple baseline and helps distinguish the value of nonlinear ensemble modeling from the contribution of the feature representation itself.

Explainability and calibration address different limitations of predictive models. Explainability concerns why a model produces a particular output, whereas calibration concerns whether a numerical probability corresponds reasonably to the observed frequency of the event. SHAP provides a unified feature-attribution approach for global and individual explanations. Platt scaling provides a post-processing transformation that can improve the reliability of classifier probabilities. The two components are therefore complementary: a model may require calibration for meaningful risk scoring while still requiring explanations for human review.

Adaptive risk prioritization is relevant when an organization has limited investigation capacity. A fixed probability threshold produces alerts whenever a score exceeds a chosen value, whereas a relative policy can select a specified fraction of the highest-risk observations within each time window. The latter can control alert volume more directly, although it may reduce recall if the selected capacity is restrictive. The adaptive experiment in this study is therefore interpreted as a policy comparison rather than as a universally superior classifier.

## 3. Objectives and Hypotheses

Building on the motivation outlined above, this study pursues four specific research objectives. First, to evaluate whether chronologically partitioned data and historical, leakage-conscious features improve the reliability of fraud-risk estimates relative to naive feature construction. Second, to compare the discrimination and threshold-dependent performance of Logistic Regression, Random Forest, Clean XGBoost, and LightGBM for transaction-level fraud detection under highly imbalanced data conditions. Third, to assess whether post-hoc probability calibration (Platt scaling) and explainability analysis (SHAP) can improve the practical usability of the strongest-performing model without degrading its ranking performance. Fourth, to examine whether an adaptive, time-windowed risk-scoring policy offers a favourable operational trade-off relative to a fixed decision threshold.

Corresponding to these objectives, the study tests four hypotheses. H1: Ensemble and gradient-boosting models (Random Forest, Clean XGBoost, LightGBM) achieve higher F1-score and PR-AUC than a Logistic Regression baseline. H2: Platt calibration reduces the Brier score of the best-performing model without changing its PR-AUC or ROC-AUC, since calibration adjusts probability values rather than rank order. H3: SHAP-based global and local explanations identify a small subset of transaction and behavioral features (e.g., transaction timing, destination-account history, and transfer type) as the dominant drivers of predicted fraud risk. H4: An adaptive top-percentile risk-scoring policy reduces the false-positive count relative to a fixed threshold, at the cost of reduced recall.

## 4. Methodological Issues

**4.1 Dataset and scope.** The experiments use the PaySim dataset, a synthetic mobile-money dataset developed for fraud-detection research (Lopez-Rojas et al., 2016). The complete dataset contains 6,362,620 transactions and 8,213 fraud cases, corresponding to an overall fraud rate of approximately 0.129%. The experimental subset retained CASH_OUT and TRANSFER transactions, producing 2,770,409 observations while retaining all 8,213 fraud cases.

Although PaySim is synthetic, its transaction structure was designed to approximate mobile-money operations comparable to services widely used in Bangladesh, including cash-in, cash-out, and peer-to-peer transfer flows analogous to bKash- and Nagad-style mobile financial services (MFS) (Lopez-Rojas et al., 2016). This structural similarity, together with PaySim's prior use in mobile-payment fraud research relevant to emerging-market contexts (Hajek et al., 2023), motivates its use here as a controlled benchmark for evaluating a fraud-detection methodology intended for eventual application to Bangladeshi MFS data. The dataset therefore functions as a proxy for transaction typology and class-imbalance characteristics rather than as a substitute for locally observed fraud behavior, and all findings are interpreted accordingly.

**4.2 Chronological split.** To preserve temporal order, transactions were divided by the step variable: steps 1–520 formed the training set (2,653,729 transactions; 5,781 fraud cases), steps 521–632 formed the validation set (78,709; 1,188 fraud cases), and steps 633–743 formed the final test set (37,971; 1,244 fraud cases). Validation data were used for threshold selection and probability calibration; test data were held out for final evaluation.

**4.3 Leakage-conscious features.** Transactions were stably sorted by step, preserving original order among observations with the same step. The final feature set contained ten variables: step, hour, day, is_night, amount, type_transfer, origin_prev_count, dest_prev_count, origin_prev_avg_amount, and amount_to_origin_avg. Historical counts and averages were constructed sequentially from preceding observations. For origin_prev_avg_amount, the current transaction amount was excluded from the cumulative calculation.

**4.4 Models and evaluation.** Logistic Regression, Random Forest, Clean XGBoost, and LightGBM were evaluated. Model discrimination was assessed with PR-AUC and ROC-AUC. Threshold-dependent performance was evaluated using precision, recall, F1-score, false-positive rate, and confusion-matrix counts. Thresholds were selected on validation data and then applied unchanged to the test set.

**4.5 Calibration.** Platt scaling was fitted to Clean XGBoost validation probabilities using Logistic Regression (Platt, 1999). No test labels were used to fit the calibrator. Calibration was assessed using the Brier score.

**4.6 Explainability.** SHAP analysis was conducted on a 5,000-observation test sample. Mean absolute SHAP values were used for global feature importance, and a local explanation was generated for a fraudulent transaction with a high predicted fraud probability.

**4.7 Adaptive risk scoring.** The adaptive policy ranked transactions by predicted risk within each 24-step window and selected the highest-risk 1% as alerts. Its results were compared with the fixed-threshold Clean XGBoost configuration.

*Table 1. Chronological data split*

| Split | Steps | Transactions | Fraud cases |
|---|---|---|---|
| Training | 1–520 | 2,653,729 | 5,781 |
| Validation | 521–632 | 78,709 | 1,188 |
| Test | 633–743 | 37,971 | 1,244 |

*Source: Author's computation using the PaySim dataset (Lopez-Rojas et al., 2016).*

The data were partitioned according to the sequential step variable rather than randomly. The training period consisted of steps 1–520, validation covered steps 521–632, and the final test period covered steps 633–743. This ordering ensures that model-development decisions are based on observations preceding the final evaluation period. Validation data were used for threshold selection and probability calibration, while the test period was reserved for final performance reporting.

The feature design combines directly observed transaction variables with sequential behavioral summaries. Time-derived variables represent the position of each transaction in the simulated sequence. Historical account variables were updated sequentially, so the current transaction did not contribute to its own historical average. This distinction is important because including the current amount in a feature described as historical would introduce information that is not available before the prediction is made. The resulting feature set therefore emphasizes information that could be constructed from preceding transaction history within the benchmark.

Four model families were evaluated using the same final feature representation: Logistic Regression, Random Forest, Clean XGBoost, and LightGBM. The comparison was designed to expose both ranking quality and threshold-dependent behavior. PR-AUC and ROC-AUC were calculated from continuous risk scores, while precision, recall, F1-score, false-positive rate, and confusion-matrix counts were calculated after applying the threshold selected on validation data. This separation prevents the final test labels from being used to optimize the reported operating point.

The validation-selected thresholds carried forward to the final test set were 0.90 for Logistic Regression, 0.93 for Random Forest, 0.96 for Clean XGBoost, and 0.99 for LightGBM. These thresholds define reproducible operating points chosen before held-out testing.

Platt scaling was applied to Clean XGBoost probabilities. A logistic calibration model was fitted using validation probabilities and labels, after which the transformation was applied to validation and test predictions. The test set was not used to fit the calibrator. Calibration quality was assessed using the Brier score, which evaluates the squared difference between predicted probability and binary outcome. Lower values indicate better probabilistic accuracy under this measure.

SHAP analysis was performed on 5,000 test observations. Global importance was summarized using mean absolute SHAP values, which measure the average magnitude of each feature's contribution without regard to direction. A local explanation was also generated for a fraudulent test observation with a very high predicted risk. The local result illustrates model attribution for one observation and should not be interpreted as a causal explanation of why the real-world event occurred.

The adaptive risk-scoring experiment divided the test sequence into 24-step windows and selected the highest-risk 1% of transactions within each window. This produces a relative ranking policy whose alert capacity is constrained by the selected proportion. It was compared directly with the fixed 0.96 Clean XGBoost threshold. The comparison therefore makes explicit the trade-off between maintaining a broader set of alerts and reducing the number of false alerts.

## 5. Results and Discussion

Experimental pipeline overview. This completed study used chronological data preparation and splitting, leakage-conscious feature engineering, comparative evaluation of four machine-learning models, validation-based threshold selection, final held-out test evaluation, Platt probability calibration, calibration analysis, SHAP global and local explainability analysis, adaptive top-1% time-window risk scoring, and reproducibility configuration with saved models and outputs.

**5.1 Comparative model performance.** Clean XGBoost achieved the strongest overall F1-score (0.5369), PR-AUC (0.5658), ROC-AUC (0.9068), and recall (0.4236) among the evaluated uncalibrated models. At its validation-selected threshold of 0.96, it produced 527 true positives and 192 false positives on the final test set. Random Forest produced the next-highest F1-score (0.4366), while LightGBM achieved 0.4350 and Logistic Regression 0.3901.

The Platt-calibrated Clean XGBoost model retained the same PR-AUC and ROC-AUC because calibration did not alter ranking. At the validation-selected calibrated threshold of 0.283, precision increased from 0.7330 to 0.8263 and the false-positive rate decreased from 0.00523 to 0.00270, while recall declined from 0.4236 to 0.3786 and F1 declined to 0.5193. Thus, within this evaluation, calibration combined with a different operating threshold shifted the model toward fewer false alerts and higher precision.

**5.2 Calibration.** Platt scaling substantially reduced the Brier score. On the validation set, the score decreased from 0.092349 to 0.011585, an improvement of 87.46%. On the test set, it decreased from 0.114086 to 0.024514, an improvement of 78.51%. These results indicate improved probability calibration according to the Brier score.

**5.3 Explainability.** The global SHAP analysis ranked step as the most influential feature, with a mean absolute SHAP value of 1.719055, followed by dest_prev_count (1.212063), type_transfer (0.711257), amount (0.711067), hour (0.651863), and day (0.627718). The remaining historical origin features had negligible mean absolute SHAP values in this trained model. The local explanation concerned a true fraud observation with predicted fraud probability 0.999813. These attributions describe the behavior of the trained model and should not be interpreted as universal causal determinants of fraud.

**5.4 Adaptive risk scoring.** The fixed-threshold Clean XGBoost policy achieved precision of 0.7330, recall of 0.4236, F1 of 0.5369, and FPR of 0.00523, with 192 false positives and 527 true positives. The adaptive top-1% policy reduced false positives to 117 and FPR to 0.00319, a reduction of approximately 39.1% in false positives. However, recall declined to 0.2122 and true positives to 264. The adaptive policy therefore reduced alert volume at the cost of detecting fewer fraudulent observations.

Overall, the experiments show that model ranking, threshold choice, probability calibration, and alert-selection policy can lead to materially different outcomes. The strongest configuration depends on the evaluation objective. If maximizing F1 within this experiment is the primary objective, the fixed-threshold Clean XGBoost model performed best. If reducing false positives is prioritized, the calibrated configuration or adaptive policy provides alternative operating points, with corresponding reductions in recall.

*Table 2. Final test-set model performance*

| Model | Threshold | PR-AUC | ROC-AUC | Precision | Recall | F1 | FPR |
|---|---|---|---|---|---|---|---|
| Clean XGBoost | 0.960 | 0.5658 | 0.9068 | 0.7330 | 0.4236 | 0.5369 | 0.00523 |
| Platt-Calibrated Clean XGBoost | 0.283 | 0.5658 | 0.9068 | 0.8263 | 0.3786 | 0.5193 | 0.00270 |
| Random Forest | 0.930 | 0.3997 | 0.8795 | 0.7023 | 0.3167 | 0.4366 | 0.00455 |
| LightGBM | 0.990 | 0.4509 | 0.8884 | 0.5281 | 0.3698 | 0.4350 | 0.01119 |
| Logistic Regression | 0.900 | 0.3212 | 0.8312 | 0.4519 | 0.3432 | 0.3901 | 0.01410 |

*Source: Author's computation using the PaySim dataset (Lopez-Rojas et al., 2016).*

*Table 3. Probability calibration results*

| Dataset | Uncalibrated Brier | Calibrated Brier | Improvement |
|---|---|---|---|
| Validation | 0.092349 | 0.011585 | 87.46% |
| Test | 0.114086 | 0.024514 | 78.51% |

*Source: Author's computation using the PaySim dataset (Lopez-Rojas et al., 2016).*

*Table 4. SHAP global feature importance*

| Rank | Feature | Mean absolute SHAP |
|---|---|---|
| 1 | step | 1.719055 |
| 2 | dest_prev_count | 1.212063 |
| 3 | type_transfer | 0.711257 |
| 4 | amount | 0.711067 |
| 5 | hour | 0.651863 |
| 6 | day | 0.627718 |
| 7 | is_night | 0.013522 |

*Source: Author's computation using the PaySim dataset (Lopez-Rojas et al., 2016).*

*Table 5. Fixed versus adaptive risk scoring*

| Method | Policy | Precision | Recall | F1 | FPR | FP | TP |
|---|---|---|---|---|---|---|---|
| Fixed Clean XGBoost | Threshold 0.960 | 0.7330 | 0.4236 | 0.5369 | 0.00523 | 192 | 527 |
| Adaptive | Top 1% / 24-step window | 0.6929 | 0.2122 | 0.3249 | 0.00319 | 117 | 264 |

*Source: Author's computation using the PaySim dataset (Lopez-Rojas et al., 2016).*

These comparative results (Table 2) establish the fixed-threshold Clean XGBoost configuration as the strongest F1-oriented operating point among the four evaluated models.

The comparison also shows that ranking and threshold performance should be reported together: the nonlinear ensemble models captured more useful structure than Logistic Regression in this benchmark. The result, however, is conditional on the PaySim data, transaction subset, chronological split, and engineered feature set, and should not be read as evidence that Clean XGBoost will necessarily outperform the alternatives on operational Bangladesh banking data.

As shown in Table 3, calibration did not alter the ranking of observations, confirming that Platt scaling improved probability quality independently of PR-AUC or ROC-AUC. At its selected threshold, the calibrated model shifted the operating point toward higher precision and fewer false alerts, at the cost of recall.

As Table 4 shows, temporal and destination-account features dominated the SHAP ranking, while the historical origin-account features contributed negligibly. The dominance of step is important to interpret cautiously, since the variable reflects progression through the simulated sequence rather than a transferable causal risk factor for fraud in Bangladesh.

The local SHAP analysis further demonstrated transaction-level attribution: the selected fraudulent observation had a predicted fraud probability of 0.999813 within a 170-case fraud sample, illustrating how an investigator-facing explanation could accompany a high-risk score. This attribution describes the model's internal reasoning rather than establishing causation, and real-world deployment would require additional validation of explanation stability and usefulness to investigators.

As Table 5 summarizes, the adaptive top-1% policy reduced false positives by approximately 39.1% relative to the fixed threshold, but at a corresponding cost in recall and true positives. This demonstrates a clear capacity-versus-detection trade-off rather than an overall improvement in classification quality.

The experiments demonstrate that there is no single metric that completely defines the quality of a fraud-risk system. Clean XGBoost was the strongest configuration when the objective was F1 at the validation-selected fixed threshold. Platt calibration was preferable when probability reliability and a higher-precision operating point were emphasized. The adaptive policy was useful for studying a stricter alert-capacity constraint, but its reduced recall shows the cost of limiting alerts to the top 1% within each 24-step window. SHAP then provided interpretability for the same modeling pipeline. Together, these results provide a more complete picture than a model leaderboard alone.

### 5.5 Visual Summary of Results

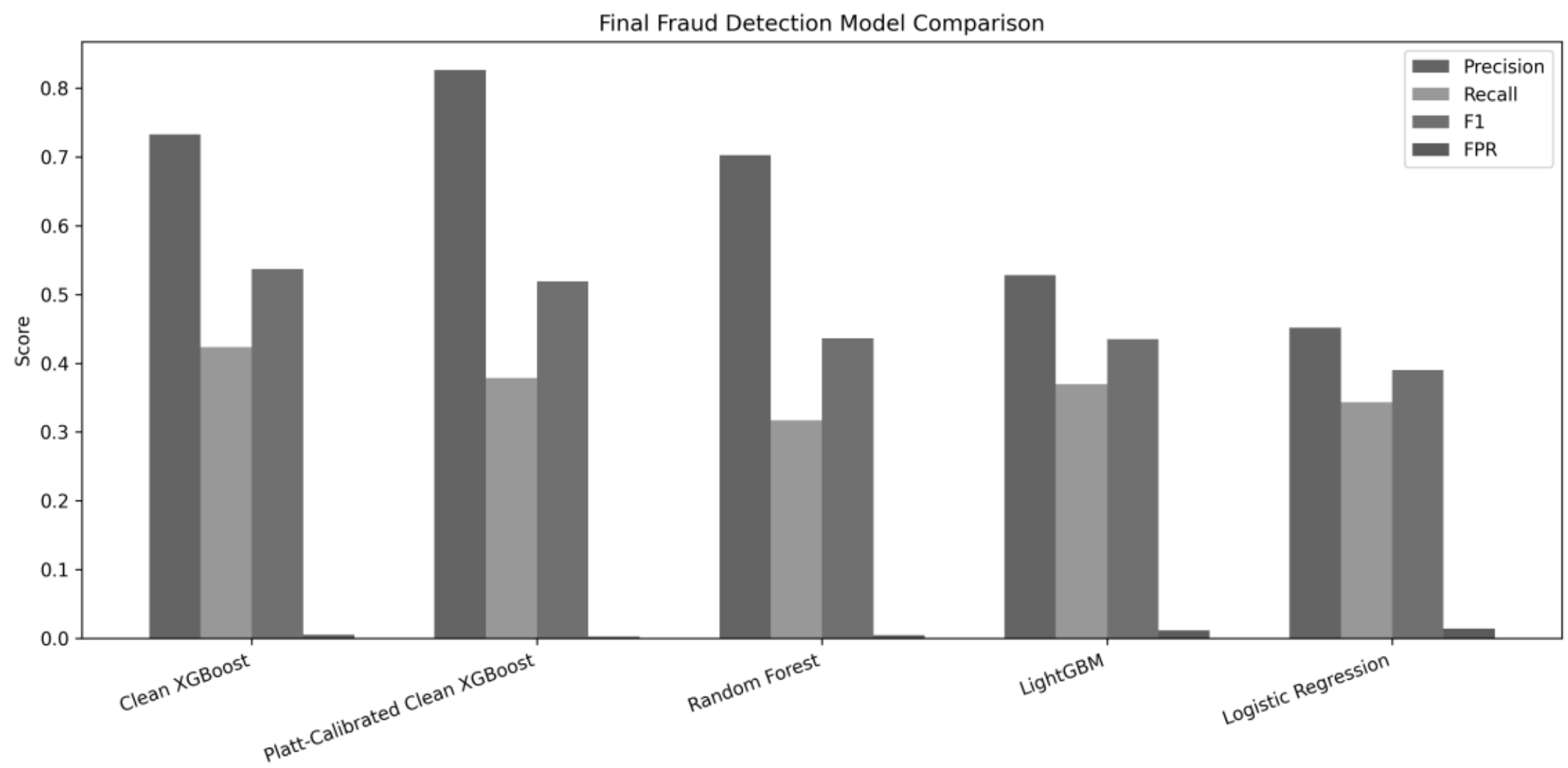

*Figure 1. Comparative model performance.*

*Source: Author's computation using the PaySim dataset (Lopez-Rojas et al., 2016).*

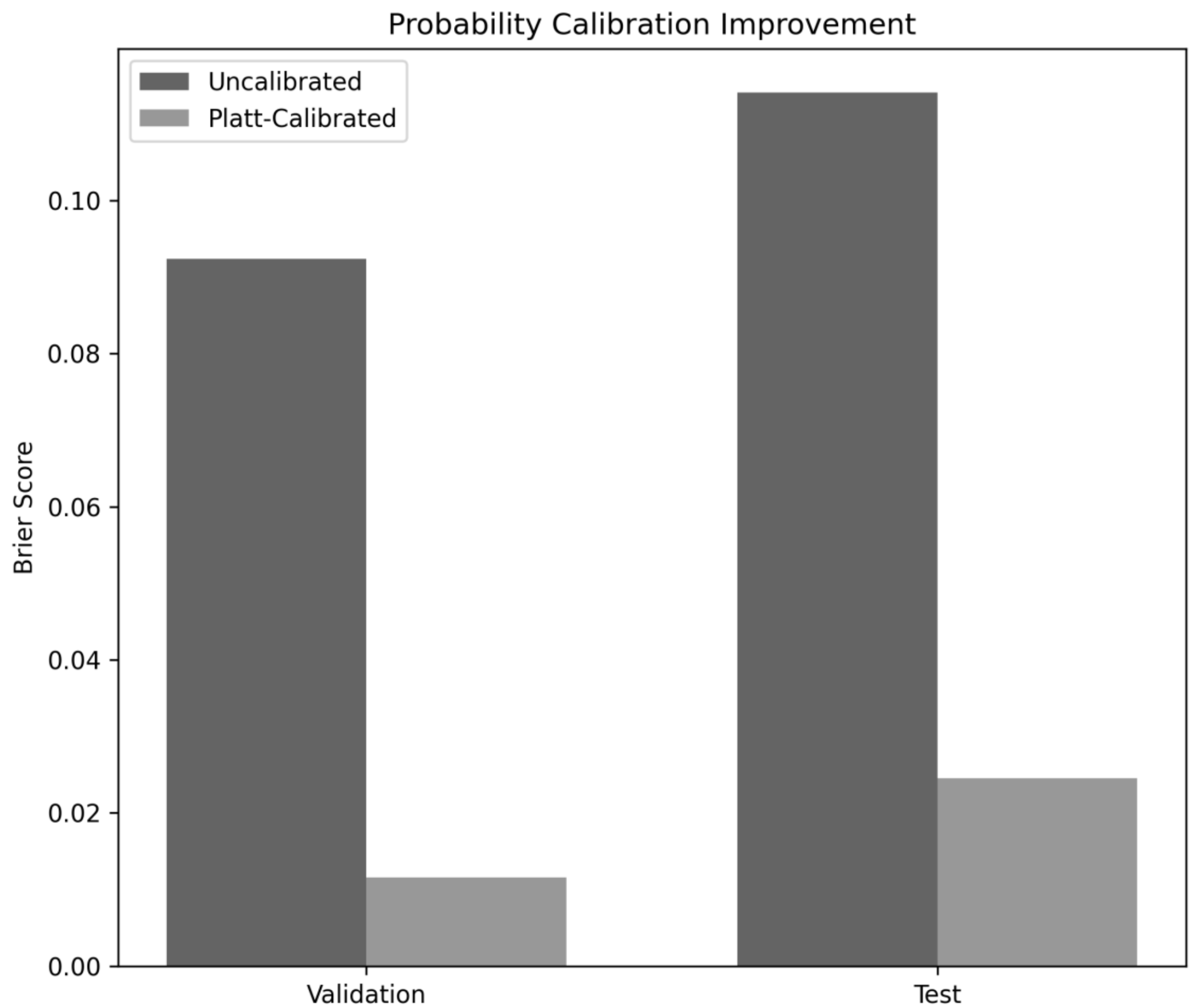

*Figure 2. Probability calibration comparison using Brier scores.*

*Source: Author's computation using the PaySim dataset (Lopez-Rojas et al., 2016).*

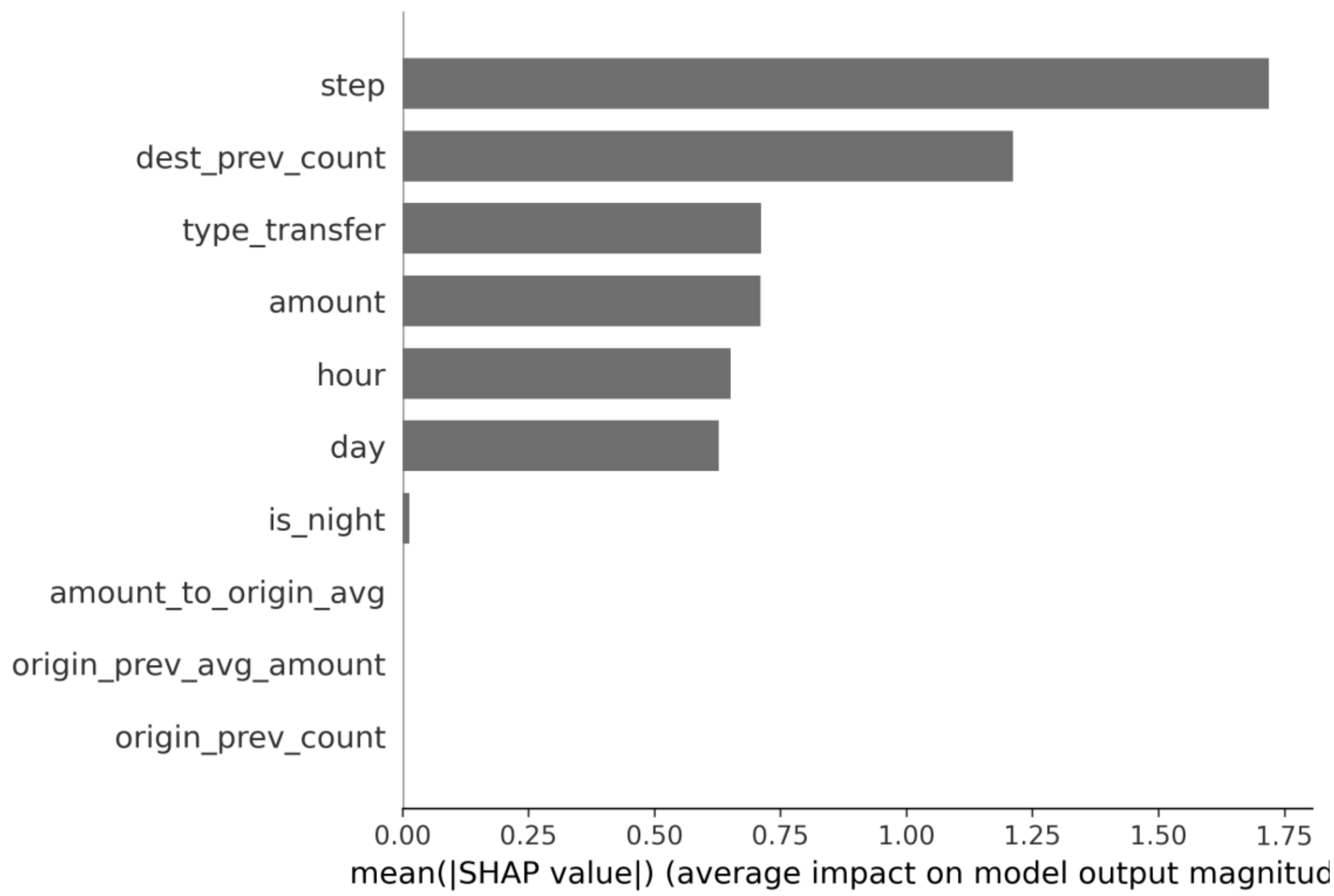

*Figure 3. SHAP global feature importance.*

*Source: Author's computation using the PaySim dataset (Lopez-Rojas et al., 2016).*

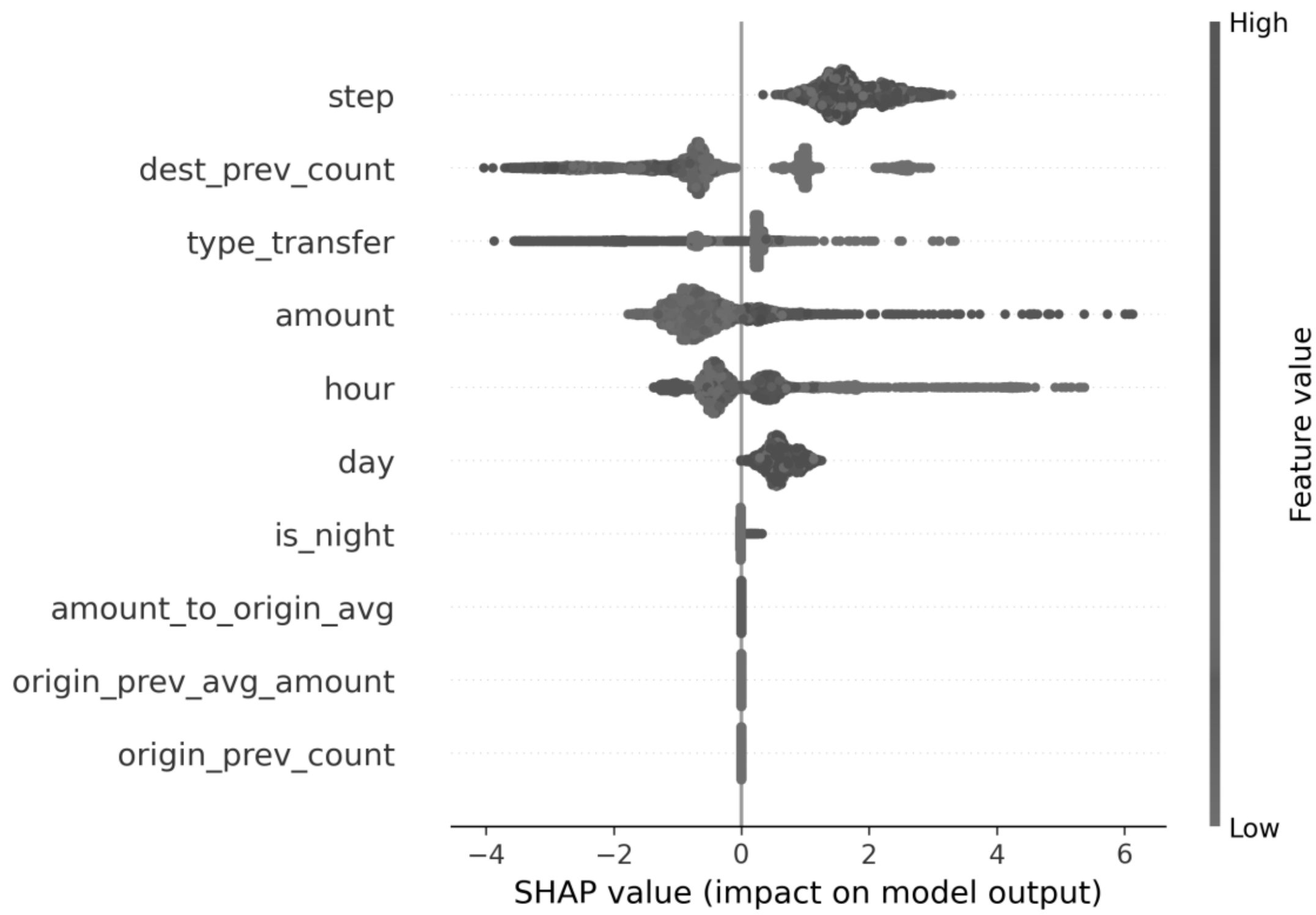

*Figure 4. SHAP summary plot.*

*Source: Author's computation using the PaySim dataset (Lopez-Rojas et al., 2016).*

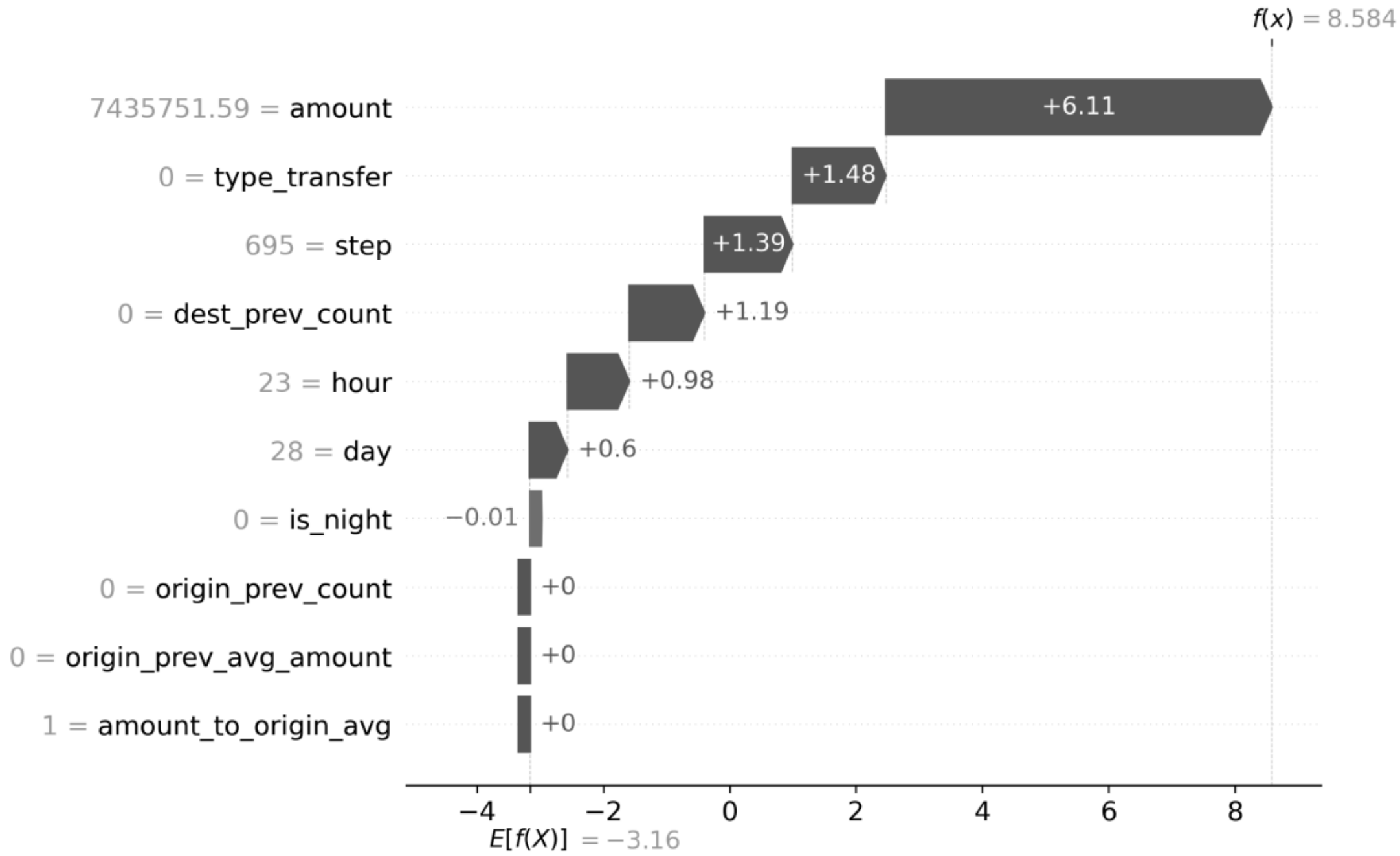

*Figure 5. Local SHAP explanation for a selected fraudulent transaction.*

*Source: Author's computation using the PaySim dataset (Lopez-Rojas et al., 2016).*

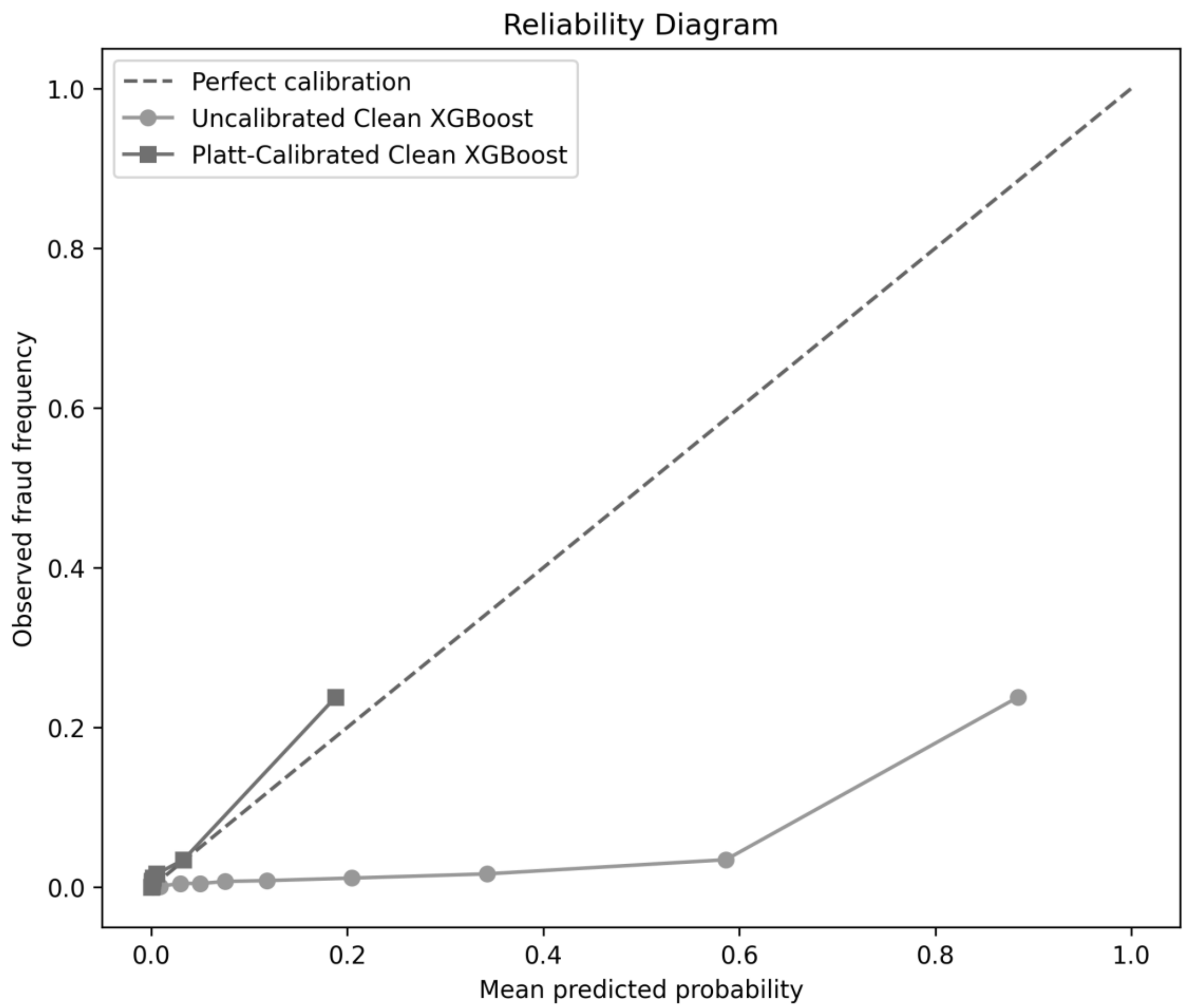

*Figure 6. Reliability diagram comparing probability estimates.*

*Source: Author's computation using the PaySim dataset (Lopez-Rojas et al., 2016).*

## 6. Findings and Policy Implications

The principal implication of this study is methodological: fraud-risk modeling can be evaluated as a sequence of connected decisions rather than as a single classifier-selection problem. Discrimination measures determine how well scores rank observations; threshold selection determines which observations become alerts; calibration determines how meaningful numerical probabilities are; SHAP reveals feature contributions; and adaptive ranking makes investigation capacity explicit. Separating these roles helps avoid treating a strong ROC-AUC or F1-score as proof that a model is ready for operational deployment.

For Bangladesh's banking sector specifically, three policy-relevant implications follow. First, Bangladesh Bank's e-KYC and fraud-monitoring guidance could encourage chronologically ordered validation, rather than random train-test splits, when institutions report fraud-detection performance, since random splits can overstate real-world accuracy. Second, calibrated probabilities combined with SHAP explanations offer a practical way to meet emerging AI-transparency and auditability expectations, supporting compliance documentation and human review of flagged transactions. Third, adaptive, capacity-constrained alert policies, rather than static thresholds, give banks and MFS operators a concrete mechanism for managing fraud-investigation workloads as digital transaction volumes grow. Realizing these implications operationally would require piloting the framework on governed, real transaction data within a specific institution.

For the Bangladesh context, the framework serves as a methodological prototype rather than a validated system: it is motivated by the country's expanding digital-finance environment, but the empirical benchmark is PaySim, not a Bangladeshi banking or MFS transaction database. The reported precision, recall, SHAP rankings, thresholds, and adaptive-policy results should not be read as direct estimates of fraud prevalence or deployed performance in Bangladesh. Operational adoption would require validation on governed local data, review of feature availability, privacy and regulatory clearance, drift monitoring, and human oversight of alerts.

PaySim is a synthetic mobile-money simulator. Its value in this study is reproducibility and controlled evaluation, but synthetic transactions cannot fully represent customer behavior, institutional controls, authentication mechanisms, fraud strategies, and data-quality issues of a specific financial institution. In addition, the experimental subset retained only CASH_OUT and TRANSFER transactions, so results should not be generalized automatically to all transaction types.

The feature set is also deliberately limited. The explainability analysis assigned the highest mean absolute SHAP value (1.719) to the step variable, which tracks sequential time progression within the synthetic PaySim benchmark. Live deployment would require replacing this explicit sequence index with cyclic temporal encodings (e.g., sine/cosine transforms of hour and day-of-week) and rolling historical account aggregates, to avoid overfitting to synthetic timeline artifacts while retaining temporal signal. Additional behavioral, device, and network features could also alter predictive performance, subject to careful temporal-availability checks to prevent data leakage.

Finally, the 1% alert proportion and 24-step window are experimental choices rather than universally optimal operational settings. Different investigation capacities, window lengths, cost assumptions, and customer-impact constraints will produce different trade-offs. A future deployment-oriented study should optimize the policy against explicit operational costs rather than adopting the top-1% rule as a default.

The completed project retained all dataset splits, model artifacts, calibration outputs, and figures to support reconstruction and independent auditing of the reported experiment.

## 7. Conclusion

This study presented a reproducible framework for explainable and adaptive machine-learning-based transaction-level fraud-risk detection, combining chronological evaluation, leakage-conscious features, comparative modeling, probability calibration, SHAP explainability, and adaptive risk prioritization.

Clean XGBoost achieved the strongest fixed-threshold test performance (F1 = 0.5369, PR-AUC = 0.5658, ROC-AUC = 0.9068). Platt scaling reduced the test Brier score from 0.114086 to 0.024514, improving probability reliability without altering ranking performance. SHAP analysis identified step, dest_prev_count, type_transfer, and amount as the most influential features, while the adaptive top-1% policy reduced false positives from 192 to 117 at the cost of lower recall (0.4236 to 0.2122), illustrating a clear capacity-versus-detection trade-off.

The contribution is therefore an integrated and reproducible experimental framework rather than a validated production system. Findings should be interpreted within the limits of the synthetic PaySim benchmark and not as direct evidence of performance on Bangladesh banking or MFS transactions. Future research should pursue external validation, additional behavioral signals, model drift monitoring, and appropriately governed real-world financial data.

More broadly, the evaluation protocol and explainability tools developed here offer a practical template that Bangladeshi banks and MFS providers can apply to their own transaction data, supporting Bangladesh Bank's evolving e-KYC and AI-governance expectations.